# Catalyzing Next-Generation Artificial Intelligence through NeuroAI

By Anthony Zador*, Sean Escola*, Blake Richards, Bence Ölveczky, Yoshua Bengio, Kwabena Boahen, Matthew Botvinick, Dmitri Chklovskii, Anne Churchland, Claudia Clopath, James DiCarlo, Surya Ganguli, Jeff Hawkins, Konrad Körding, Alexei Koulakov, Yann LeCun, Timothy Lillicrap, Adam Marblestone, Bruno Olshausen, Alexandre Pouget, Cristina Savin, Terrence Sejnowski, Eero Simoncelli, Sara Solla, David Sussillo, Andreas S. Tolias, Doris Tsao

* Equal contributions

**Abstract**: *Neuroscience has long been an essential driver of progress in artificial intelligence (AI). We propose that to accelerate progress in AI, we must invest in fundamental research in NeuroAI. A core component of this is the embodied Turing test, which challenges AI animal models to interact with the sensorimotor world at skill levels akin to their living counterparts. The embodied Turing test shifts the focus from those capabilities like game playing and language that are especially well-developed or uniquely human to those capabilities – inherited from over 500 million years of evolution – that are shared with all animals. Building models that can pass the embodied Turing test will provide a roadmap for the next generation of AI.*

Over the coming decades, Artificial Intelligence (AI) will transform society and the world economy in ways that are as profound as the computer revolution of the last half century and likely at an even faster pace. This AI revolution presents tremendous opportunities to unleash human creativity and catalyze economic growth, relieving workers from performing the most dangerous and menial jobs. However, to reach this potential, we still require advances that will make AI more human-like in its capabilities. Historically, neuroscience has been a critical driver and source of inspiration for improvements in AI, particularly those that made AI more proficient in areas that humans and other animals excel at, such as vision, reward-based learning, interacting with the physical world, and language[1,2]. It can still play this role. To accelerate progress in AI and realize its vast potential, we must invest in fundamental research in "NeuroAI."

The seeds of the current AI revolution were planted decades ago, mainly by researchers attempting to understand how brains compute[3]. Indeed, the earliest efforts to build an "artificial brain" led to the invention of the modern "von Neumann computer architecture," for which John von Neumann explicitly drew upon the very limited knowledge of the brain available to him in the 1940s[4,5]. Later, the Nobel-prize winning work of David Hubel and Torsten Wiesel on visual processing circuits in the cat neocortex inspired the deep convolutional networks that have catalyzed the recent revolution in modern AI[6–8]. Similarly, the development of reinforcement learning was directly inspired by insights into animal behavior and neural activity during learning[9–15]. Now, decades later, applications of ANNs and RL are coming so quickly that many observers assume that the long-elusive goal of human-level intelligence – sometimes referred to as "artificial general intelligence" – is within our grasp. However, in contrast to the optimism of those outside the field, many front-line AI researchers believe that major breakthroughs are needed before we can build artificial systems capable of doing all that a human, or even a much simpler animal like a mouse, can do.

Although AI systems can easily defeat any human opponent in games such as chess[16] and Go[17], they are not robust and often struggle when faced with novel situations. Moreover, we have yet to build effective systems that can walk to the shelf, take down the chess set, set up the pieces, and move them around during a game, although recent progress is encouraging[18]. Similarly, no machine can build a nest, forage for berries, or care for young. Today's AI systems cannot compete with the sensorimotor capabilities of a four-year old child or even simple animals. Many basic capacities required to navigate new situations – capacities that animals have or acquire effortlessly – turn out to be deceptively challenging for AI, partly because AI systems lack even the basic abilities to interact with an unpredictable world. A growing number of AI researchers doubt that merely scaling up current approaches will overcome these limitations. Given the need to achieve more natural intelligence in AI, it is quite likely that new inspiration from naturally intelligent systems is needed[19].

Historically, many key AI advances, such as convolutional ANNs and reinforcement learning, were inspired by neuroscience. Neuroscience continues to provide guidance – e.g., attention-based neural networks were loosely inspired by attention mechanisms in the brain[20–23] – but this is often based on findings that are decades old. The fact that such cross-pollination between AI and neuroscience is far less common than in the past represents a missed opportunity. Over the last decades, through efforts such as the NIH BRAIN initiative and others, we have amassed an enormous amount of knowledge about the brain. ***The emerging field of NeuroAI, at the intersection of neuroscience and AI, is based on the premise that a better understanding of neural computation will reveal fundamental ingredients of intelligence and catalyze the next revolution in AI.*** This will eventually lead to artificial agents with capabilities that match those of humans. The NeuroAI program we advocate is driven by the recognition that AI historically owes much to neuroscience and the promise that AI will continue to learn from it–but only if there is a large enough community of researchers fluent in both domains. We believe the time is right for a large-scale effort to identify and understand the principles of biological intelligence and abstract those for application in computer and robotic systems.

It is tempting to focus on the most characteristically human aspects of intelligent behavior, such as abstract thought and reasoning. However, the basic ingredients of intelligence – adaptability, flexibility, and the ability to make general inferences from sparse observations – are already present in some form in basic sensorimotor circuits which have been evolving for hundreds of millions of years. As AI pioneer Hans Moravec[24] put it, abstract thought "is a new trick, perhaps less than 100 thousand years old….effective only because it is supported by this much older and much more powerful, though usually unconscious, sensorimotor knowledge." This implies that the bulk of the work in developing general AI can be achieved by building systems that match the perceptual and motor abilities of animals and that the subsequent step to human-level intelligence would be considerably smaller. This is good news because progress on the first goal can rely on the favored subjects of neuroscience research – rats, mice, and non-human primates – for which extensive and rapidly expanding behavioral and neural datasets can guide the way. Thus, we believe that the NeuroAI path will lead to necessary advances if we figure out the core capabilities that all animals possess in embodied sensorimotor interaction with the world.

## NeuroAI Grand Challenge: The Embodied Turing Test

In 1950, Alan Turing proposed the "imitation game" as a test of a machine's ability to exhibit intelligent behavior indistinguishable from that of a human[25] (Figure 1, left). In that game, now known as the Turing test, a human judge evaluates natural language conversations between a real human and a machine trained to mimic human responses. By focusing on conversational abilities, Turing evaded asking whether a machine could "think," a question he considered impossible to answer. The Turing test is based on the implicit belief that language represents the pinnacle of human intelligence and that a machine capable of conversation must surely be intelligent.

Until recently, no artificial system could come close to passing the Turing test. However, a class of modern AI systems called "large language models" can now engage in surprisingly cogent conversations[26]. In part, their success reveals how easily we can be tricked into imputing intelligence, agency, and even consciousness to our interlocutor[27]. Impressive though these systems are, because they are not grounded in real-world experiences, they nonetheless continue to struggle with many basic aspects of causal reasoning and physical common sense. Thus, the Turing test does not probe our prodigious perceptual and motor abilities to interact with and reason about the physical world, abilities shared with animals and honed through countless generations of natural selection.

We therefore propose an expanded "embodied Turing test," one that includes advanced sensorimotor abilities (Figure 1, right). The spirit of the original Turing test was to establish a simple qualitative standard against which our progress toward building artificially intelligent machines can be judged. This embodied Turing test would benchmark and compare the interactions with the world of artificial systems versus humans and other animals. Similar ideas have been proposed previously[28–32]. However, in light of recent advances enabling large-scale behavioral and neural measurements, as well as large-scale simulations of embodied agents *in silico,* we believe the time is ripe to instantiate a major research effort in this direction. Because each animal has its own unique set of abilities, each animal defines its own embodied Turing test: An artificial beaver might be tested on its ability to build a dam, and an artificial squirrel on its ability to jump through trees. Nonetheless, many core sensorimotor capabilities are shared by almost all animals, and the ability of animals to rapidly evolve the sensorimotor skills needed to adapt to new environments suggests that these core skills provide a solid foundation. This implies that after developing an AI system to faithfully reproduce the behavior of one species, the adaptation of this system to other species – and even to humans – may be straightforward. Below we highlight a few of the characteristics that are shared across species.

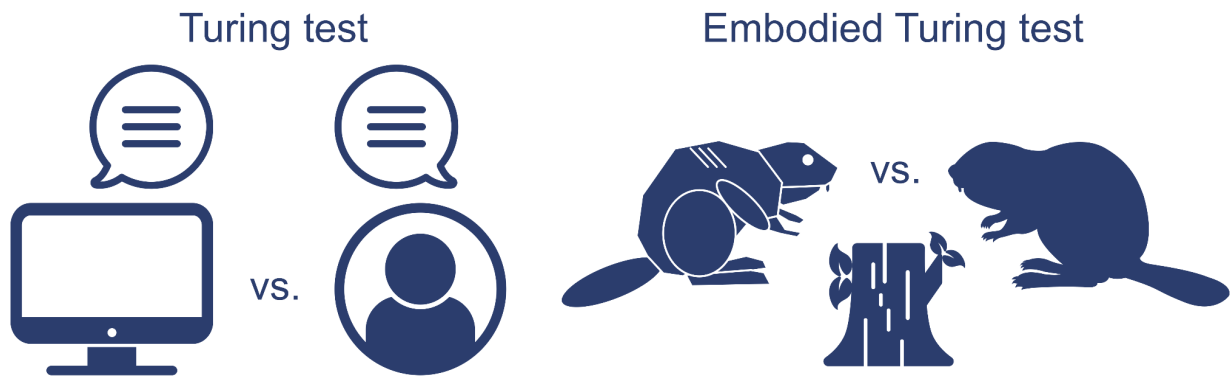

**Figure 1.** Turing tests: comparisons between the performance of AI systems and their living counterparts. *Left*: The original Turing test as proposed by Alan Turing[25]. If a human tester cannot determine whether their interlocutor is an AI system or another human, the AI passes the test. Modern large language models have made substantial progress towards passing this test[26]. *Right*: The embodied Turing test. An AI animal model – whether robotic or in simulation – passes the test if its behavior is indistinguishable from that of its living counterpart. No AI systems are close to passing this test. Here, an artificial beaver is tested on the species-specific behavior of dam construction.

**Animals engage their environments.** The defining feature of animals is their ability to move around and interact with their environment in purposeful ways. Despite recent advances in optimal control, reinforcement learning, and imitation learning, robotics is still far from achieving animal-level abilities in controlling their bodies and manipulating objects, even in simulation. Of course, neuroscience can provide guidance about the kinds of modular and hierarchical architectures that could be adapted to artificial systems to give them these capabilities[33]. It can also provide us with design principles like partial autonomy (how low-level modules in a hierarchy act semi-autonomously in the absence of input from high-level modules) and amortized control (how movements generated at first by a slow planning process are eventually transferred to a fast reflexive system). These principles could guide the design of systems for perception, action selection, locomotion, and fine-grained control of limbs, hands, and fingers. Understanding how specific neural circuits participate in different tasks could also inspire solutions for other forms of 'intelligence,' including in more cognitive realms. For example, we speculate that incorporating principles of circuitry for low-level motor control could help provide a better basis for higher-level motor planning in AI systems.

**Animals behave flexibly.** Another goal is to develop AI systems that can engage a large repertoire of flexible and diverse tasks in a manner that echoes the incredible range of behaviors that individual animals can generate. Modern AI can easily learn to outperform humans at video games like *Breakout* using nothing more than pixels on a screen and game scores[34]. However, these systems, unlike human players, are brittle and highly sensitive to small perturbations: changing the rules of the game slightly, or even a few pixels on the input, can lead to catastrophically poor performance[35]. This is because these systems learn a mapping from pixels to actions that need not involve an understanding of the agents and objects in the game and the physics that governs them. Similarly, a self-driving car does not inherently know about the danger of a crate falling off a truck in front of it unless it has literally seen examples of crates falling off trucks leading to bad outcomes. And even if it has been trained on the dangers of falling crates, the system might consider an empty plastic bag being blown out of the car in front of it as an obstacle to avoid at all costs rather than an irritant, again, because it doesn't actually understand what a plastic bag is or how unthreatening it is physically. This inability to handle scenarios that have not appeared in the training data is a significant challenge to widespread reliance on AI systems.

To be successful in an unpredictable and changing world, an agent must be flexible and master novel situations by using its general knowledge about how such situations are likely to unfold. This is arguably what animals do. Animals are born with most of the skills needed to thrive or can rapidly acquire them from limited experience, thanks to their strong foundation in real-world interaction, courtesy of evolution and development[36]. Thus, it is clear that training from scratch for a specific task is not how animals obtain their impressive skills; animals do not arrive into the world *tabula rasa* and then rely on large labeled training sets to learn. Although machine learning has been pursuing approaches for sidestepping this *tabula rasa* limitation, including self-supervised learning, transfer learning, continual learning, meta learning, one-shot learning and imitation learning[37], none of these approaches comes close to achieving the flexibility found in most animals. Thus, we argue that understanding the neural circuit-level principles that provide the foundation for behavioral flexibility in the real-world, even in simple animals, has the potential to greatly increase the flexibility and utility of AI systems. Put another way, we can greatly accelerate our search for general-purpose circuits for real-world interaction by taking advantage of the optimization process that evolution has already engaged in[38–45].

**Animals compute efficiently.** One important challenge for modern AI – that our brains have overcome – is energy efficiency. Training a neural network requires enormous amounts of energy. For example, training a large language model such as GPT-3 requires over 1000 megawatts-hours, enough to power a small town for a day[46]. Biological systems are, by contrast, much more energy efficient: The human brain uses about 20 watts[47]. The difference in energy requirement between brains and computers derives from differences in information processing. First, at an algorithmic level, modern large-scale ANNs, such as large language models[26], rely on very large feedforward architectures with self-attention to process sequences over time[23], ignoring the potential power of recurrence for processing sequential information. One reason for this is that currently we do not have efficient mechanisms for credit assignment calculations in recurrent networks. In contrast, brains utilize flexible recurrent architectures that can solve the temporal credit assignment problem with great efficiency. Uncovering the mechanisms by which this happens could potentially enable us to increase the energy efficiency of artificial systems. Alternatively, it has been proposed that the synaptic dynamics within adjacent dendritic spines could serve as a mechanism for learning sequential structure, a scheme that could potentially be efficiently implemented in hardware[48]. Second, at an implementation level, neural circuits differ from digital computers. Neural circuits compute effectively despite the presence of unreliable or "noisy" components. For example, synaptic release, the primary means of communication between neurons, can be so unreliable that only one in every ten messages is transmitted[49]. Furthermore, neurons interact mainly by transmitting action potentials (spikes), an asynchronous communication protocol. Like the interactions between conventional digital elements, the output of a neuron can be viewed as a string of 0s and 1s; but unlike a digital computer, the energy cost of a "1" (i.e. of a spike) is several orders of magnitude higher than that of a "0"[50]. Because biological circuits operate in a regime where spikes are sparse – even very active neurons rarely fire at more than 100 spikes per second and typical cortical firing rates may be less than 1 spike/second – they are much more energy efficient[51]. Spike-based computation has also been shown to be orders of magnitude faster and more energy efficient in recent hardware implementation[52].

**A roadmap for solving the embodied Turing test**

How might artificial systems that pass the embodied Turing test be developed? One natural approach would be to do so incrementally, guided by our evolutionary history. For example, almost all animals engage in goal-directed locomotion; they move toward some stimuli (e.g. food sources) and away from others (e.g. threats). Layered on top of these foundational abilities are more sophisticated skills, such the ability to combine different streams of sensory information (e.g. visual and olfactory), to use this sensory information to distinguish food sources and threats, to navigate to previous locations, to weigh possible rewards and threats to achieve goals, and to interact with the world in precise ways in service of these goals. Most of these – and many other – sophisticated abilities are found to some extent in even very simple organisms, such as worms. In more complex animals, such as fish and mammals, these abilities are elaborated and combined with new strategies to enable more powerful behavioral strategies.

This evolutionary perspective suggests a strategy for passing the embodied Turing test by breaking it down into a series of incrementally challenging ones that build on each other, and iteratively optimizing on this series[53]. Specifically, the embodied Turing test comprises challenges that include a wide range of organisms used in neuroscience research, including worms, flies, fish, rodents and primates. This would enable us to deploy the vast amount of knowledge we have accumulated about the behavior, biomechanics, and neural mechanisms of these model organisms to both precisely define each species-specific embodied Turing test and serve as strong inductive biases to guide the development of robust AI controllers that can pass it.

The performance of these artificial agents could be compared with that of animals. Rich behavioral datasets representing a large swath of a species' ethogram have now been collected and can be deployed to benchmark performance on species-specific embodied Turing tests. Furthermore, these datasets are being rapidly expanded given new tools in 3D videography[54–57]. Additionally, detailed biomechanical measurements support highly realistic animal body models, complete with skeletal constraints, muscles, tendons, and paw features[58]. Combined with the open-sourcing of powerful, fast physics simulators and virtual environments[59,60], these models will afford the opportunity for embodied Turing test research to be performed *in silico* at scale[33]. Finally, existing extensive neural datasets with simultaneous neural recordings across multiple brain regions during behavior, combined with increasingly detailed neural anatomy and connectomics, provide a powerful roadmap for the design of AI systems that can control virtual animals to recapitulate the behaviors of their *in vivo* counterparts and thus pass the embodied Turing test.

Importantly, the specifics of the embodied Turing test for each species can be tuned to the needs of different groups of researchers. We can test the capacity of AI systems in terms of sensorimotor control, self-supervised and continual learning, generalization, memory-guided behavior on both short and life-long timescales, and social interactions. Despite these potentially different areas of interest, the challenges that compose the embodied Turing test can be standardized to permit the quantification of progress and comparison between research efforts. Standardization can be fostered by stakeholders including government and private funders, large research organizations such as the Allen Institute, and major collaborations like the International Brain Lab, with an eye toward the development of common APIs and support for competitions as has been an important impetus for much progress in machine learning and robotics[61,62]. Ultimately, virtual organisms that demonstrate successful recapitulation of behaviors of interest can be adapted to the physical world with additional efforts in robotics and deployed to solve real-world problems.

**What we need**

Achieving these goals will require significant resources deployed in three main areas. ***First, we must train a new generation of AI researchers who are equally at home in engineering/computational science and neuroscience***. These researchers will chart fundamentally new directions in AI research by drawing on decades of progress in neuroscience. The greatest challenge will be in determining how to exploit the synergies and overlaps in neuroscience, computational science, and other relevant fields to advance our quest: identifying what details of the brain's circuitry, biophysics, and chemistry are important and what can be disregarded in the application to AI. There is thus a critical need for researchers with dual training in AI and neuroscience to apply insights from neuroscience to advance AI and to help design experiments that generate new insights relevant to AI. Although there is already some research of this type, it exists largely at the margins of mainstream neuroscience; training in neuroscience has thus far been motivated and funded mainly by the goal of improving human health and of understanding the brain as such. This lack of alignment between fields might explain, *e.g.*, the multi-decade gap between Hubel and Wiesel's discovery of the structure of the visual system[6] and the development and application of convolutional neural networks in modern machine learning[8]. Thus, the success of a NeuroAI research program depends on the formation of a community of researchers for whom the *raison d'être* of their training is to exploit synergies between neuroscience and AI. Explicit design of new training programs can ensure that the NeuroAI research community reflects the demographics of society as a whole and is equipped with the ethical tools needed to ensure that the development of AI benefits society[63].

***Second, we must create a shared platform capable of developing and testing these virtual agents.*** One of the greatest technical challenges that we will face in creating an iterative, embodied Turing test and evolving artificial organisms to pass it is the amount of computational power required. Currently, training just one large neural network model on a single embodied task (e.g. control of a body in 3-dimensional space) can take days on specialized distributed hardware[64]. For multiple research groups to iteratively work together to optimize and evaluate a large number of agents over multiple generations on increasingly complex embodied Turing tests, a large investment in a shared computational platform will be required. Much like a particle accelerator in physics or large telescope in astronomy, this sort of large-scale shared resource will be essential for moving the brain-inspired AI research agenda forward. It will require a major organizational effort, with government and ideally also industry support, that has as its central goal scientific progress on animal and human-like intelligence.

Third, *we must support fundamental theoretical and experimental research on neural computation*. We have learned a tremendous amount about the brain over the last decades, through the efforts of the NIH, in no small measure due to the BRAIN Initiative, and other major funders, and we are now reaching an understanding of the vast diversity of the brain's individual cellular elements, neurons, and how they function as parts of simple circuits. With these building blocks in place, we are poised to shift our focus toward understanding how the brain functions as an integrated intelligent system. This will require insight into how a hundred billion neurons of a thousand different types, each one communicating with thousands of other neurons, with variable, adaptable connections, are wired together, and the computational capabilities – the intelligence – that emerges. We must *reverse engineer the brain* to abstract the underlying principles. Taking advantage of the powerful synergies between neuroscience and AI will require program and infrastructure support to organize and enable research across the disciplines at a large scale.

Fortunately, there is now broad political agreement that investments in AI research are essential to humanity's technological future. Indeed, IARPA (Intelligence Advanced Research Projects Activity) was a pioneer in this field, launching the Machine Intelligence from Cortical Networks (MICrONS) project. This project spearheaded the collection of an unprecedented data set consisting of a portion of a mouse connectome and associated functional responses with the specific goal of catalyzing the development of next-generation AI algorithms[65]. Nonetheless, community-wide efforts to bridge the fields of neuroscience and AI will require robust investments from government resources, as well as oversight of project milestones, commercialization support, ethics, and big bets on innovative ideas. In the U.S., there are currently some lines of federal resourcing, such as the NSF's National Artificial Intelligence Research Institutes, explicitly dedicated to driving innovation and discovery in AI from neuroscience research, but these are largely designed to support a traditional academic model with different groups investigating different questions, rather than the creation of a centralized effort that could create something like the embodied Turing test. Likewise, AI support grants in the U.S. are predominantly ancillary programs through the NIH, NSF, DoD, and even the EPA – each of which have their own directives and goals – and this pattern is shared by funding agencies globally. This leaves a significant funding gap for technology development as an end in itself. The creation of overarching directives either through existing entities, or as a stand-alone agency, to support NeuroAI and AI research would drive this mission and consequently unlock the potential for AI to benefit humanity.

## Conclusions

Despite the long history of neuroscience driving advances in AI and the tremendous potential for future advances, most engineers and computational scientists in the field are unaware of the history and opportunities. The influence of neuroscience on shaping the thinking of von Neumann, Turing and other giants of computational theory are rarely mentioned in a typical computer science curriculum. Leading AI conferences such as NeurIPS, which once served to showcase the latest advances in both computational neuroscience and machine learning, now focus almost exclusively on the latter. Even some researchers aware of the historical importance of neuroscience in shaping the field often argue that it has lost its relevance. "Engineers don't study birds to build better planes" is the usual refrain. But the analogy fails, in part because pioneers of aviation did indeed study birds[66,67], and some still do[68,69]. Moreover, the analogy fails also at a more fundamental level: The goal of modern aeronautical engineering is not to achieve "bird-level" flight, whereas a major goal of AI is indeed to achieve (or exceed) "human-level" intelligence. Just as computers exceed humans in many respects, such as the ability to compute prime numbers, so too do planes exceed birds in characteristics such as speed, range and cargo capacity. But if the goal of aeronautical engineers were indeed to build a machine with the "bird-level" ability to fly through dense forest foliage and alight gently on a branch, they would be well-advised to pay very close attention to how birds do it. Similarly, if AI aims to achieve animal-level common-sense sensorimotor intelligence, researchers would be well-advised to learn from animals and the solutions they evolved to behave in an unpredictable world.

## Acknowledgements

This Perspective grew out of discussions following the first conference From Neuroscience to Artificially Intelligent Systems (NAISys), held in CSHL in 2020, and includes many of the invited speakers from that meeting. A.Z. would like to thank Cat Donaldson for convincing him that this manuscript would be worth writing, and for contributing to early versions. B.Ö. would like to acknowledge valuable contributions on the manuscript from Diego Aldorando. The authors would like to acknowledge the following funding sources – A.Z.: Schmidt Foundation, Eleanor Schwartz Foundation, Robert Lourie Foundation; S.E.: NIH 5U19NS104649, 5R01NS105349; R.B.: CIFAR, NSERC RGPIN-2020-05105, RGPAS-2020-00031; B.Ö.: NIH R01NS099323, R01GM136972; Y.B.: CIFAR; K.B.: Stanford Institute for Human-Centered Artificial Intelligence, NSF 2223827; A.C.: NIH U19NS123716; J.D.: Semiconductor Research Corporation, DARPA, Office of Naval Research MURI-114407, MURI-N00014-21-1-2801, N00014-20-1-2589, NSF CCF-1231216, NSF 2124136, Simons SCGB 542965; S.G.: Simons Foundation, James S. McDonnell Foundation, NSF CAREER Award; A.M.: Federation of American Scientists; B.O.: NSF IIS-1718991; C.S.: NIH 1R01MH125571-01, R01NS127122, NSF 1922658, Google faculty award; E.S.: Simons Foundation, NIH EY022428; S.S.: NIH NINDS NS053603; D.S.: Simons 543049; A.T.: IARPA D16PC00003, DARPA HR0011-18-2-0025, NIH R01EY026927, R01MH109556, NSF NeuroNex DBI-1707400; D.T.: HHMI.

## Competing interests

The authors declare their engagements with the following relevant for-profit entities – A.Z.: Cajal Neuroscience; S.E.: Herophilus; R.B.: Google DeepMind; B.Ö.: Blackbird Neuroscience; K.B.: Femtosense Inc., Radical Semiconductor, Neurovigil; C.C.: Google DeepMind; S.G.: Meta; J.H.: Numenta; K.K.: Paradromics; T.L.: Google DeepMind; A.M.: Google DeepMind, Kernel; D.S.: Meta; A.T.: Vathes Inc., Upload AI LLC, BioAvatar LLC.

## Author contributions

A.Z. conceived the project and wrote the first draft. A.Z., S.E., B.R., B.Ö. provided extensive editing to successive drafts. All other authors provided additional guidance and edits at various stages.